\documentclass[letterpaper]{article} 
\usepackage{aaai24}  
\usepackage{times}  
\usepackage{helvet}  
\usepackage{courier}  
\usepackage[hyphens]{url}  
\usepackage{graphicx} 
\usepackage{natbib}  
\usepackage{caption} 
\usepackage{algorithm}
\usepackage{algorithmic}

\usepackage{times}
\usepackage{latexsym}

\usepackage{graphicx}
\usepackage{array, makecell} %
\usepackage{caption}
\usepackage{subcaption}
\usepackage{tablefootnote}
\usepackage{footnote}
\usepackage{multirow}
\usepackage{hhline}
\usepackage{booktabs}
\usepackage{tabularx,colortbl}
\usepackage[table]{xcolor}

\usepackage{amsmath} 
\usepackage{newfloat}
\usepackage{listings}
\DeclareCaptionStyle{ruled}{labelfont=normalfont,labelsep=colon,strut=off} 
\floatstyle{ruled}
\newfloat{listing}{tb}{lst}{}
\floatname{listing}{Listing}
\title{LongFin: A Multimodal Document Understanding Model for Long Financial Domain Documents}
\author{
    Ahmed Masry,
    Amir Hajian
}
\affiliations{
    Arteria AI\\
    ahmed.masry@arteria.ai, amir.hajian@arteria.ai
}

\newcommand{\ahmed}[1]{\textcolor{blue}}
\newcommand{\dataset}[1]{LongForms}
\newcommand{\model}[1]{LongFin}
\usepackage{bibentry}

\begin{document}

\maketitle

\begin{abstract}
Document AI is a growing research field that focuses on the comprehension and extraction of information from scanned and digital documents to make everyday business operations more efficient. Numerous downstream tasks and datasets have been introduced to facilitate the training of AI models capable of parsing and extracting information from various document types such as receipts and scanned forms. Despite these advancements, both existing datasets and models fail to address critical challenges that arise in industrial contexts. Existing datasets primarily comprise short documents consisting of a single page, while existing models are constrained by a limited maximum length, often set at 512 tokens. Consequently, the practical application of these methods in financial services, where documents can span multiple pages, is severely impeded. 
To overcome these challenges, we introduce \model\\, a multimodal document AI model capable of encoding up to 4K tokens. We also propose the \dataset\\ dataset, a comprehensive financial dataset that encapsulates several industrial challenges in financial documents. Through an extensive evaluation, we demonstrate the effectiveness of the \model\\ model on the \dataset\\ dataset, surpassing the performance of existing public models while maintaining comparable results on existing single-page benchmarks.
\end{abstract}

\section{Introduction}

There has been a noticeable industrial interest surrounding the automation of data extraction from various documents, including receipts, reports, and forms to minimize manual efforts and enable seamless downstream analysis of the extracted data \cite{zhang2020rapid, layoutlm}. However, the process of parsing documents poses several challenges, including obscure information within scanned documents that may result in Optical Character Recognition (OCR) errors, complex layouts (such as tables), and intricate content structures.

To investigate and address these challenges, several datasets have been made available. These datasets encompass a wide range of tasks, such as classification \cite{rvl-cdip}, semantic entity recognition \cite{cord, funsd}, relation extraction \cite{funsd}, question answering \cite{docvqa}, and key information extraction \cite{sroie}.
Nonetheless, a significant limitation shared by these datasets is that they mostly consist of single-page documents with a limited amount of content.
As a consequence, these datasets fail to capture various challenges inherent in parsing lengthy documents spanning multiple pages, which are commonly encountered in the financial industry.
Financial reports and documents can become exceedingly lengthy, necessitating a comprehensive understanding of the entire context to effectively analyze and extract pertinent information.

\begin{figure}[t!]
\centering
    \includegraphics[width=0.40\textwidth,keepaspectratio]{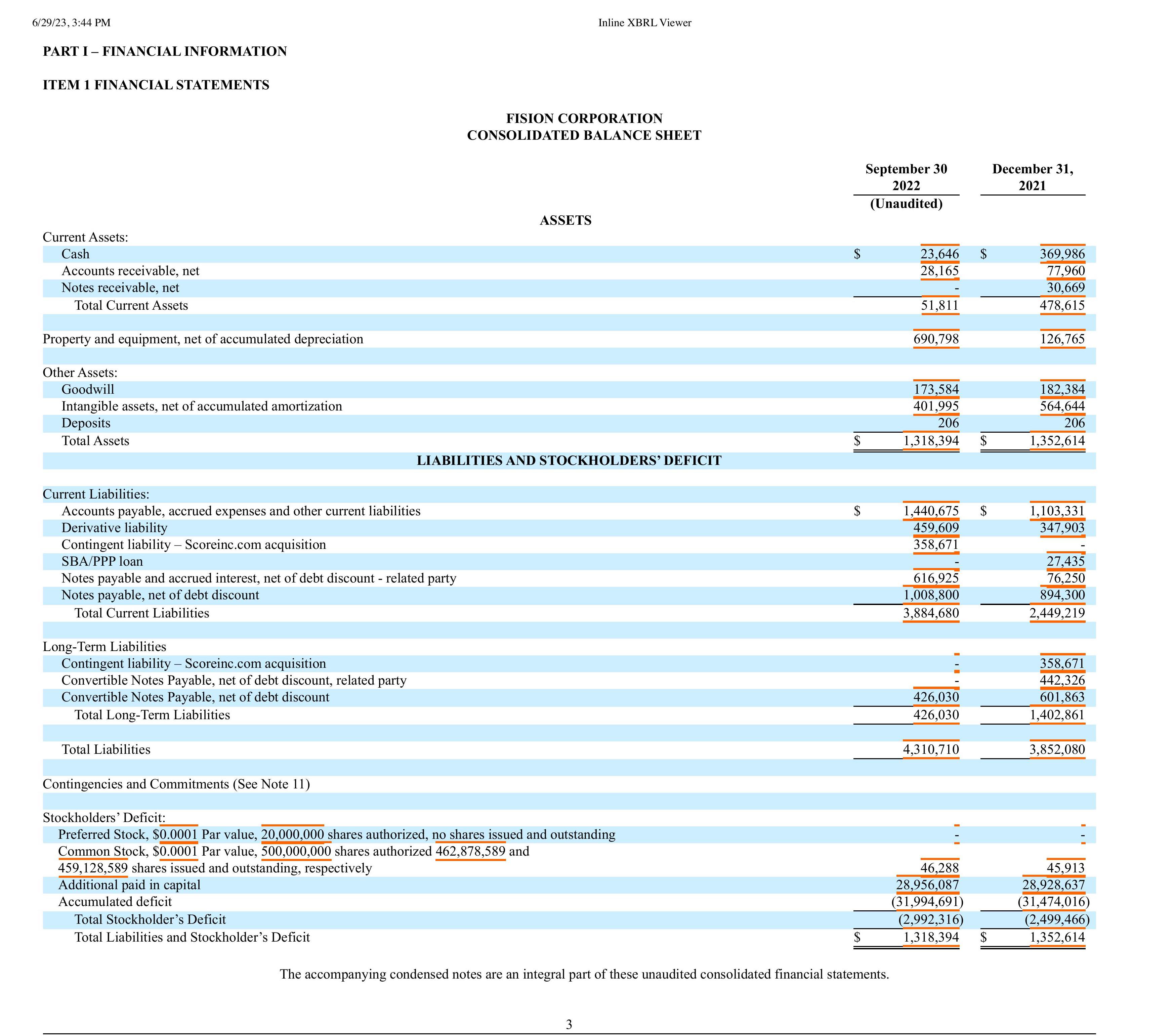}
\caption{\small  First page from a 4-page example financial form in the \dataset\\ dataset.
The information in these documents is spread over a mix of tables and text spanning multiple pages which makes it challenging for short-context models. }
\vspace{-4mm}
\label{models}
\end{figure}

The limitations inherent in existing datasets have a direct impact on the capabilities of the proposed models. In the literature, two primary lines of work have emerged: \emph{(i)} OCR-dependent architectures \cite{lilt, layoutlm, layoutlmv2, layoutlmv3, udop} \emph{(ii)} OCR-free models \cite{donut, pix2struct}. OCR-dependent models typically employ transformer-based text encoders and incorporate spatial information by leveraging the words' coordinates in the documents as additional embeddings. One notable exception is UDOP \cite{udop} which consists of an encoder-decoder architecture. Conversely, OCR-free models typically employ a vision encoder to process the scanned document image and a text decoder to generate the desired information. Nevertheless, a common limitation shared by most of these models is their design and pretraining to handle a maximum of 512 tokens or process a single input image.

In this work, we introduce two main contributions. Firstly, we present the \dataset\\ dataset, a comprehensive financial dataset primarily comprising 140 long forms where the task is formulated as named entity recognition. Due to privacy concerns and proprietary limitations, we were unable to utilize our internal resources to construct this dataset. Consequently, we obtained financial statements from the SEC website\footnote{https://www.sec.gov/edgar/}, aligning our tasks to encompass the significant challenges encountered in the financial documents which require a deep understanding of lengthy contexts.
Secondly, we propose \model\\, a multimodal document understanding model capable of processing up to 4K tokens. Our approach builds upon LiLT \cite{lilt}, one of the state-of-the-art multimodal document understanding models. Additionally, we incorporate techniques that effectively extend the capabilities of text-only models, such as RoBERTa \cite{roberta}, to handle longer sequences, as demonstrated by Longformer \cite{longformer}. By leveraging these techniques, our proposed model exhibits enhanced performance in processing lengthy financial forms. The efficacy of our approach is extensively evaluated, showcasing its effectiveness and paving the way for numerous commercial applications in this domain.

\section{Related Work}
\label{sec:relatedwork}

\subsection{Document Datasets}
Several recently released datasets in the field of document understanding have contributed significantly to advancing research in this area. The RVL-CDIP dataset \cite{rvl-cdip} introduced a classification task, encompassing 400K scanned documents categorized into 16 classes, such as forms and emails. Another notable dataset, DocVQA \cite{docvqa}, focuses on document question answering and comprises 50K question-answer pairs aligned with 12K scanned images. In addition, the CORD dataset \cite{cord} consists of 11K scanned receipts, challenging models to extract 54 different data elements (e.g., phone numbers and prices).
Furthermore, the FUNSD dataset \cite{funsd} was proposed, featuring 200 scanned forms. This dataset primarily revolves around two key tasks: semantic entity recognition (e.g., header, question, answer) and relation extraction (question-answer pairs). FUNSD is particularly relevant to our dataset, \dataset\\, as it also mainly consist of forms.  
However, FUNSD and all the above-mentioned datasets mainly focus on short contexts, as they typically consist of single-page documents. In contrast, our \dataset\\ dataset primarily consists of multi-page documents, presenting unique challenges that demand a comprehensive understanding of lengthy contexts which is common in the financial industry.

\vspace{-1mm}
\subsection{Document AI Models}
Numerous document understanding models have been developed to tackle the challenges posed by the aforementioned benchmark datasets. These models can be broadly categorized into two main groups: OCR-free and OCR-dependent models.
OCR-free models, exemplified by Donut \cite{donut} and Pix2Struct \cite{pix2struct}, typically employ vision transformer-based encoders to process input images and text decoders to handle output generation. These models are often pretrained on OCR-related tasks, enabling them to comprehend the text embedded within scanned documents effectively.
On the other hand, OCR-dependent models, including LayoutLM \cite{layoutlm}, LayoutLMv2 \cite{layoutlmv2}, LayoutLMv3 \cite{layoutlmv3}, LiLT \cite{lilt}, DocFormer \cite{docformer} and UDOP \cite{udop}, rely on external OCR tools to initially extract underlying text from scanned documents. To incorporate layout information, these models utilize specialized positional embeddings, encoding the coordinates of each word in the document. Additionally, some models, such as LayoutLMv2, LayoutLMv3, DocFormer, and UDOP, employ visual embeddings created by splitting the image into patches. These visual embeddings, along with the text and layout embeddings, are fed into the models. While LayoutLM, LayoutLMv2, LayoutLMv3, DocFormer, and LiLT adopt an encoder-only architecture, UDOP is based on the T5 model \cite{t5}, which follows an encoder-decoder architecture.
Despite the impressive achievements of these models, they share a common limitation: they are typically designed to process a single page or a maximum of 512 tokens, thereby restricting their applicability to multi-page documents. \cite{longdocument} proposed a multimodal document understanding model that can process up to 4096 tokens, however their code is not publicly available and their model performance deteriorates on the short-context datasets such as FUNSD \cite{funsd}. 
In contrast, our proposed model, \model\\, works efficiently on both short and long contexts (to up 4096 tokens), making it particularly well-suited for a variety of real-world industrial applications.

\section{\dataset\\ Dataset}
\label{sec:longfin}
Due to privacy constraints, we are unable to utilize internal documents for dataset construction. Instead, we turn to publicly available financial reports and tailor our dataset, \dataset\\, to emulate the challenges encountered in our proprietary datasets. This approach ensures the task's alignment with real-world financial contexts without violating privacy.

\subsection{Dataset Collection \& Preparation}
\label{sec:dataset_collection}
To construct \dataset\\, we leverage the EDGAR database \footnote{https://www.sec.gov/edgar/}, a comprehensive repository of financial filings and reports submitted by US companies. These filings are based on different financial form types (e.g., 10-K, 10-Q) which vary in structure and content. Our dataset primarily centers around the SEC Form 10-Q, which provides a detailed quarterly report on a company's finances. This specific form is chosen due to its similarity in both structure and content to to the documents we frequently encounter in the financial services industry. 

We download 140 10-Q forms that were published between 2018 and 2023. This deliberate decision to keep the dataset relatively small is intended to mirror the limited data challenges commonly encountered in real-world scenarios, particularly in the finance domain, where strict data confidentiality prevents access to large-scale datasets. Consequently, it is common practice to construct smaller datasets that mimic the proprietary datasets \cite{madl2023approximate}. Furthermore, our dataset size aligns with recently published datasets, such as the FUNSD dataset \cite{funsd} which primarily consists of single-page forms. Inspired by the FUNSD dataset, we perform a random split of the \dataset\\ dataset and divide the dataset into 105 training documents, which account for 75\% of the total dataset, and 35 testing documents, representing the remaining 25\%.

\subsection{Dataset Description \& Setup}
\label{sec:task_desctiption}
Our dataset, \dataset\\, is formulated as a Named Entity Recognition (NER) task. The dataset consists of $N$ examples, denoted as $D = \{d_i, w_i, b_i, n_i\}_{i=1}^N$, where $d_i$ represents a PDF document, $w_i$ represents the list of words, $b_i$ represents the list of bounding boxes, and $n_i$ represents a list of entities present in the document. To obtain the words ($w_i$) and their bounding boxes ($b_i$), each PDF document is processed using the pdftotext\footnote{https://pypi.org/project/pdftotext/} tool. Moreover, we define six entity types: \emph{(i)} Total Assets, \emph{(ii)} Cash at the beginning of the period (Beginning Cash), \emph{(iii)} Cash at the end of the period (End Cash), \emph{(iv)} Cash provided by financial activities (Financial Cash), \emph{(v)} Net change in cash (Change in Cash), and \emph{(vi)} Quarter Keys. As shown in Table \ref{tab:data_stats}, our \dataset\\ dataset contains 140 forms that consist of 685 pages, 168458 words, and 1128 entities in total. The models are trained to predict $n_i$ given both $w_i$ and $b_i$.

\begin{table}[t]
  \centering
  \small
  \scalebox{0.80}{
  \begin{tabular}{l|c|c|c|c}
  \toprule
     \makecell{Dataset \\ Split} & \makecell{\#Forms} & \makecell{\#Pages} &  \makecell{\#Words} & \makecell{\#Entities}  \\
    \midrule
    Train & 105 & 514 & 125094 & 843  \\
    Test  & 35 & 171 & 43364 & 285 \\
   \midrule
    Overall & 140 & 685 & 168458 & 1128 \\
    \bottomrule
  \end{tabular}
  }
  \vspace{-2mm}
  \caption{\dataset\\ dataset statistics.}
  \vspace{-4mm}
  \label{tab:data_stats}
\end{table}

\section{\model\\ Model}
\label{sec:longlilt}

\begin{figure*}[t!]
\begin{subfigure}[t]{.40\textwidth}
\centering
    \includegraphics[width=1\textwidth,keepaspectratio]{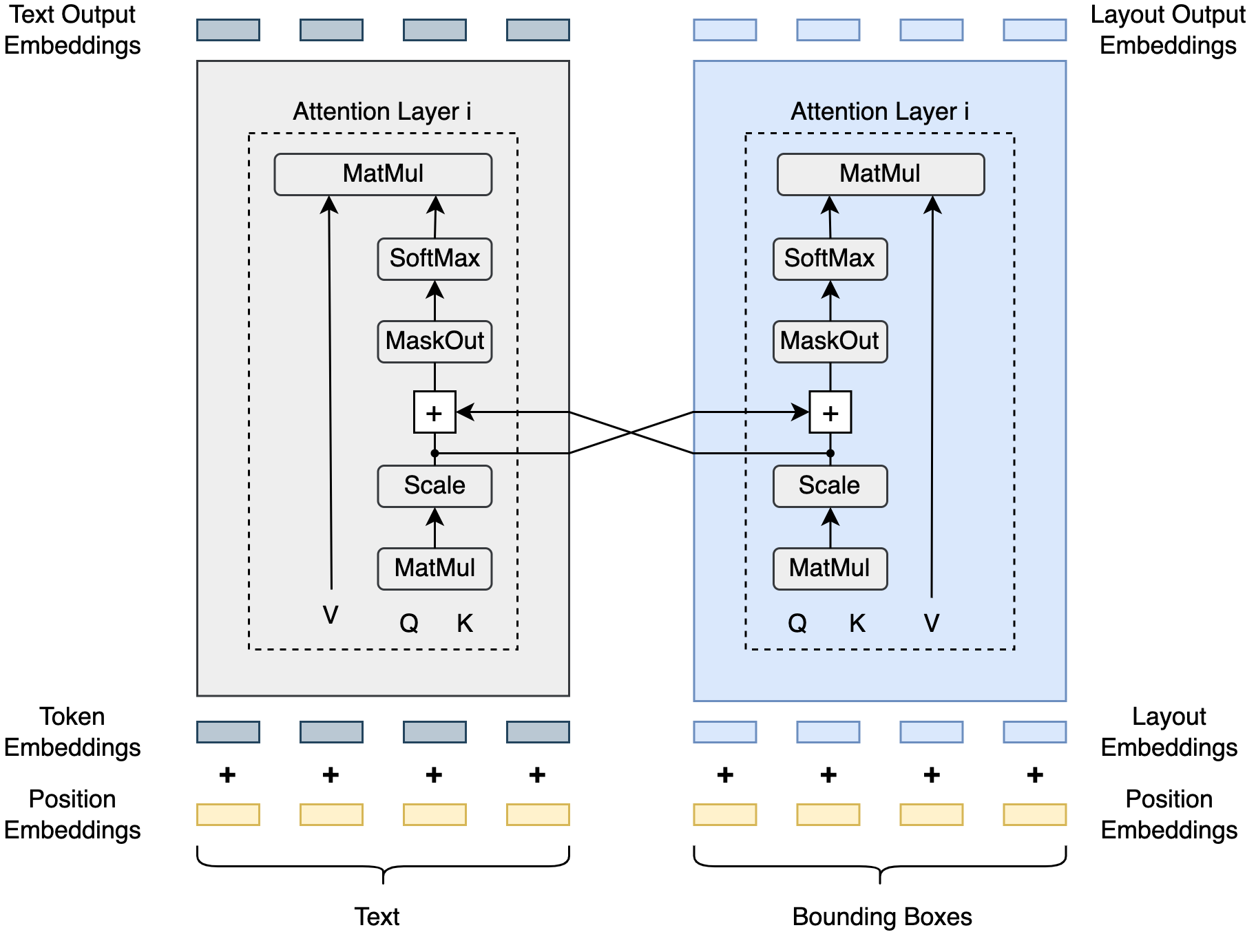}
\caption{\small {\model\\}}
\label{fig:longlilt}
\end{subfigure}
\hspace{35mm}
\begin{subfigure}[t]{.23\textwidth}
\centering
    \raisebox{8mm}{\includegraphics[width=1\textwidth,keepaspectratio]{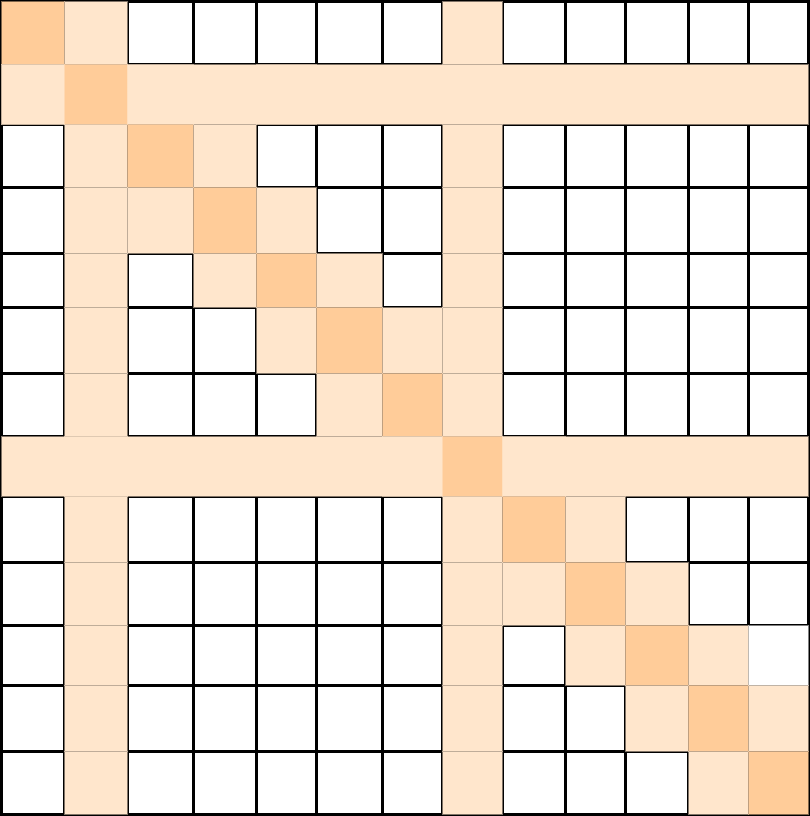}}

\caption{\small {Local + Global Atention}}
\label{fig:attention}
\end{subfigure}
\vspace{-0.8em}
\caption{\small (a) The architecture of the \model\\ model. It mainly consists of two encoders: text encoder and layout encoder which are connected through the BiACM layer. (b) A visualization of the employed local (sliding window) and global attention mechanisms to process long sequences.}
\vspace{-4mm}
\label{fig:models}
\end{figure*}

\subsection{Architecture} 
Figure \ref{fig:models} illustrates the overall architecture of our proposed model, \model\\, which builds upon recently published models: LiLT \cite{lilt} and Longformer \cite{longformer}. Similar to LiLT \cite{lilt}, \model\\ comprises three primary components: a text encoder, a layout encoder, and the BiACM (bidirectional attention complementation mechanism) layer \cite{lilt}. However, \model\\ introduces additional mechanisms, namely sliding window local attention and interval-based global attention, to effectively handle long contexts within both the text and layout encoders. One key advantage of \model\\ is its ability to scale linearly with the input sequence length, in contrast to the quadratic scaling ($O(n^2)$) observed in the original transformers' \cite{vaswani2017attention} attention mechanism. This linear scaling, inspired by the Longformer model \cite{longformer}, allows \model\\ to efficiently handle long contexts up to 4K tokens.

\vspace{-1mm}
\subsubsection{Text Encoder}
For the text encoder in \model\\, we adopt the Longformer \cite{longformer} model, which has been pretrained to handle long textual contexts of up to 4096 tokens. As depicted in Figure \ref{fig:longlilt}, the input to the text encoder consists of two types of embeddings: text embeddings ($E_{T}$) and absolute position embeddings ($E_{P}$). These embeddings are added together to produce the final embeddings ($E_{final}$). Subsequently, a layer normalization \cite{layernormalization} operation is applied, and the resulting output is fed into the encoder.

The attention mechanism in \model\\ incorporates two types of attention: local attention and global attention. The local attention employs a sliding window approach, where each token attends to the 512 local tokens surrounding it. On the other hand, the global attention involves a set of global tokens, selected at intervals of 100. While other approaches \cite{longformer, longdocument} may employ different methods for selecting global tokens, such as random selection or task-specific strategies, we limit our experimentation to interval-based selection for simplicity and due to limited computational resources. Each token in the input sequence attends to these global tokens, in addition to its local context as shown in Figure \ref{fig:attention}. This combination of local and global attention mechanisms enhances the model's ability to capture both local context and broader global dependencies within the long input sequences.

\subsubsection{Layout Encoder}
For the layout encoder in \model\\, we adopt the layout encoder utilized in the LiLT model \cite{lilt}. Similar to the text encoder, the input for the layout encoder comprises two types of embeddings: absolute position embeddings and layout embeddings. Each word in the input document is associated with a bounding box that defines its location within the document layout. This bounding box is represented by four numbers: $x_0$, $y_0$, $x_1$, and $y_1$, which correspond to the coordinates of the top-left and bottom-right points of the bounding box. To normalize these coordinates within the range [0,1000], we use the page's height and width. 

To generate the layout embedding for each word, each coordinate in the normalized bounding box is used to obtain an embedding vector. The different coordinates' embedding vectors are then concatenated and projected using a linear layer. The resulting layout embeddings are added to the absolute position embeddings to obtain the final embeddings. These final embeddings are then fed into the layout encoder. Similar to the text encoder, we also employ the local \& global attention mechanisms in the layout encoder to process long sequences. 

\subsubsection{BiACM}
To facilitate communication between the text encoder and layout encoder, we incorporate the BiACM layer from the LiLT model \cite{lilt}. As depicted in Figure \ref{fig:longlilt}, the BiACM layer adds the scores resulting from the multiplication of keys and queries from both encoders. In LiLT, a detach operation is applied to the scores generated by the text encoder before passing them to the layout encoder. This detachment prevents the layout encoder from backpropagating into the text encoder during pretraining, promoting better generalization when fine-tuning the model with different language text encoders. However, since our focus is primarily on the English language for our applications, we have chosen to remove the detach operation to expedite pretraining, given our limited computational resources.

\vspace{-1mm}
\subsection{Pretraining}
\label{sec:pretraining}
To pretrain \model\\, we utilize the IIT-CDIP \cite{iit} dataset which contains 11M scanned images that make up 6M documents. We obtain the OCR annotations (words and their bounding boxes) from OCR-IDL \cite{ocraws} which used the AWS Textract API\footnote{https://aws.amazon.com/textract/}. We initialize our text encoder from Longformer \cite{longformer} and our layout encoder from LiLT \cite{lilt} layout encoder. Since the LiLT layout encoder was pretrained on inputs with a maximum length of 512 tokens, we copy LiLT's pretrained positional embeddings eight times to initialize our layout encoder positional embeddings, which consist of 4096 embedding vectors. This enables the layout encoder to handle longer sequences while leveraging the pretrained positional information from the LiLT model.

\begin{figure}[t!]
\centering
    \includegraphics[width=0.45\textwidth,keepaspectratio]{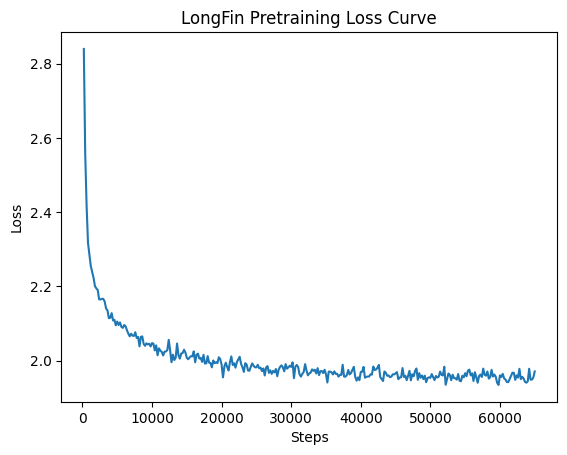}
\caption{\small  \model\\ pretraining loss curve. The loss starts at 2.84 and oscillated between 1.97 and 1.94 near convergence. }
\vspace{-4mm}
\label{fig:loss_curve}
\end{figure}

For the pretraining of \model\\, we employ the Masked Visual-Language Modeling task \cite{bert, lilt}. In this task, 15\% of the tokens in the input to the text encoder are masked. In 80\% of the cases, we replace the masked tokens with the \textsc{[MASK]} token. In 10\% of the cases, we replace the masked tokens with random tokens. In the remaining 10\%, we keep the original token unchanged.
Inspired by Longformer \cite{longformer}, we pretrain the model for 65K steps with a learning rate of 3e-5 and batch size of 12 on one A100 GPU. We set the warmup steps to 500 and use the AdaFactor optimizer \cite{shazeer2018adafactor}. Also, we utilize gradient checkpointing \cite{gradientcheckpointing} to enable using a large batch size. The pretraining loss curve is shown in Figure \ref{fig:loss_curve}

\section{Experiments \& Evaluation}
\label{sec:evaluation}

\vspace{-1mm}
\subsection{Tasks \& Datasets}
To assess the generalizability of \model\\ on both short and long contexts, we evaluate \model\\ on two existing short (single-page) datasets: FUNSD \cite{funsd} and CORD \cite{cord} to show the generalizability of our model on short contexts as well as our newly created \dataset\\ dataset. 

\noindent \textbf{$\bullet$ {\sc{FUNSD}}}: This dataset comprises 200 scanned forms and requires models to extract four main entities: headers, questions, answers, and other relevant information. Additionally, it involves linking questions with their corresponding answers, thereby encompassing named entity recognition and relation extraction tasks. We mainly focus on the named entity recognition task and use the entity-level F1 score as our evaluation metric.

\noindent \textbf{$\bullet$ {\sc{CORD}}}: With over 11,000 receipts, this dataset focuses on extracting 54 different data elements (e.g., phone numbers) from receipts. The task can be formulated as named entity recognition or token classification. For evaluation, we use the entity-level F1 score.

\vspace{-2mm}
\subsection{Baselines}

To demonstrate the effectiveness of \model\\ on our \dataset\\ dataset, we compare it against a set of publicly available text and text+layout baselines that are capable of handling both short and long input sequences.
For the text baselines, we select the following models: \emph{(i)} BERT \cite{bert} which is a widely used text-based model known for its strong performance on short context tasks (512 tokens), \emph{(ii)} Longformer \cite{longformer} which is specifically designed to handle text long texts (up to 4096 tokens). For the text+layout baseline, we utilize LiLT \cite{lilt}, which is one of the state-of-the-art models for document understanding \footnote{LayoutLMv3 \cite{layoutlmv3} is another state-of-the-art document understanding model, but its usage is limited to non-commercial applications}. For the short context models, we split the \dataset\\ documents into chunks that can fit within 512 tokens. Table \ref{tab:finetuningdetails}  shows the hyperparameters of the different models when finetuning on the \dataset\\ dataset. It also presents the hyperparameters we used when finetuning \model\\ on the previous single-page datasets. All the finetuning experiments were performed on one A100 and one T4 GPUs. 

\subsection{Results}

\begin{table}[t!]
\centering
\scalebox{0.64}{\begin{tabular}{l|c|cc}
\toprule
Model & Modalities & \textbf{FUNSD} & \textbf{CORD} \\
 & & \small ($F1$) & \multicolumn{1}{c}{\small ($F1$)} \\
\hline
\multicolumn{4}{c}{\textbf{Short Context Models (512 tokens)}} \\
$\text{BERT}_{BASE}$  & T &  60.26 &  89.68 \\
$\text{RoBERTa}_{BASE}$  & T  & 66.48 & 93.54 \\
$\text{LayoutLM}_{BASE}$  & T+L & 79.27 & 94.72 \\
$\text{LayoutLMv2}_{BASE}$  & T+L+V & 82.76 &  94.95 \\
$\text{LayoutLMv3}_{BASE}$  & T+L+V & \underline{90.29} & \underline{96.56} \\
$\text{DocFormer}_{BASE}$  & T+L+V & 83.34 & 96.33 \\
$\text{LiLT}_{BASE}$ & T+L & 88.41 & 96.07  \\

\midrule
\multicolumn{4}{c}{\textbf{Long Context Models (4096 tokens)}} \\
\rowcolor{gray!20}  $\text{Longformer}_{BASE}$  & T & 71.4 & 90.41  \\
\rowcolor{gray!20}  \cite{longdocument}  & T+L & 77.1 & --\tablefootnote{The code of \cite{longdocument} is not publicly available.}  \\
\rowcolor{gray!20} $\text{\model\\}_{BASE}$  \textcolor{blue}{(ours)} & T+L  & \underline{87.03} & \underline{94.81}  \\
\bottomrule
\end{tabular}
}
\vspace{-2mm}
\caption{
    \small Accuracy of the different models on FUNSD and CORD datasets. The second column shows the modalities used by each model where T refers to Text, L refers to Layout, and V refers to Vision.}
\vspace{-3mm}
\label{tab:prev_datasets}
\end{table}

\subsection{Previous (Single-Page) Datasets}
As shown in Table \ref{tab:prev_datasets}, \model\\ outperforms other long-context models such as Longformer \cite{longformer} and \cite{longdocument} on the previous datasets that mainly consist of single-page documents.
The performance disparity is particularly pronounced on the FUNSD dataset \cite{funsd}, where all documents have very short textual content (less than 512 tokens). 
Notably, \model\\ also achieves comparable performance to the short-context models on these datasets. 
This comparison highlights the superior generalization ability of our model, \model\\, which performs well on both short and long contexts. In contrast, the performance of \cite{longdocument} model deteriorates on short-context documents. 

\begin{table}[t!]
\centering
\scalebox{0.64}{\begin{tabular}{l|c|ccc}
\toprule
Model & Modalities & \textbf{Precision} & \textbf{Recall} & \textbf{F1} \\
\hline
\multicolumn{5}{c}{\textbf{Short Context Models}} \\
$\text{BERT}_{BASE}$  & T  & 35.82 & 30.18 & 32.76 \\
$\text{LiLT}_{BASE}$ & T+L & 35.55 & 43.24 & 39.02 \\

\midrule
\multicolumn{5}{c}{\textbf{Long Context Models}} \\
\rowcolor{gray!20}  $\text{Longformer}_{BASE}$  & T & \textbf{49.50} & 45.20 & 47.25 \\
\rowcolor{gray!20} $\text{\model\\}_{BASE}$  \textcolor{blue}{(ours)} & T+L  & 47.67 & \textbf{56.16} & \textbf{51.57} \\
\bottomrule
\end{tabular}
}
\vspace{-2mm}
\caption{
    \small Results of the different models on \dataset\\. We evaluate using the entity-level F1 score.}
\vspace{-4mm}
\label{tab:longfin_results}
\end{table}

\subsection{\dataset\\ Dataset}
As presented in Table \ref{tab:longfin_results}, the performance results on our \dataset\\ dataset highlight the advantage of our model, \model\\, compared to the short-context models. This observation emphasizes the significance of long-context understanding when working with financial documents. There is also a noticeable difference in performance between the text models (BERT \cite{bert} and Longformer \cite{longformer}) and text+layout models (LiLT \cite{lilt} and \model\\). This is mainly because the documents in \dataset\\ contain diverse layouts that might be challenging for text-only models.

\begin{figure*}[t!]
\centering
    \includegraphics[width=0.60\textwidth,keepaspectratio]{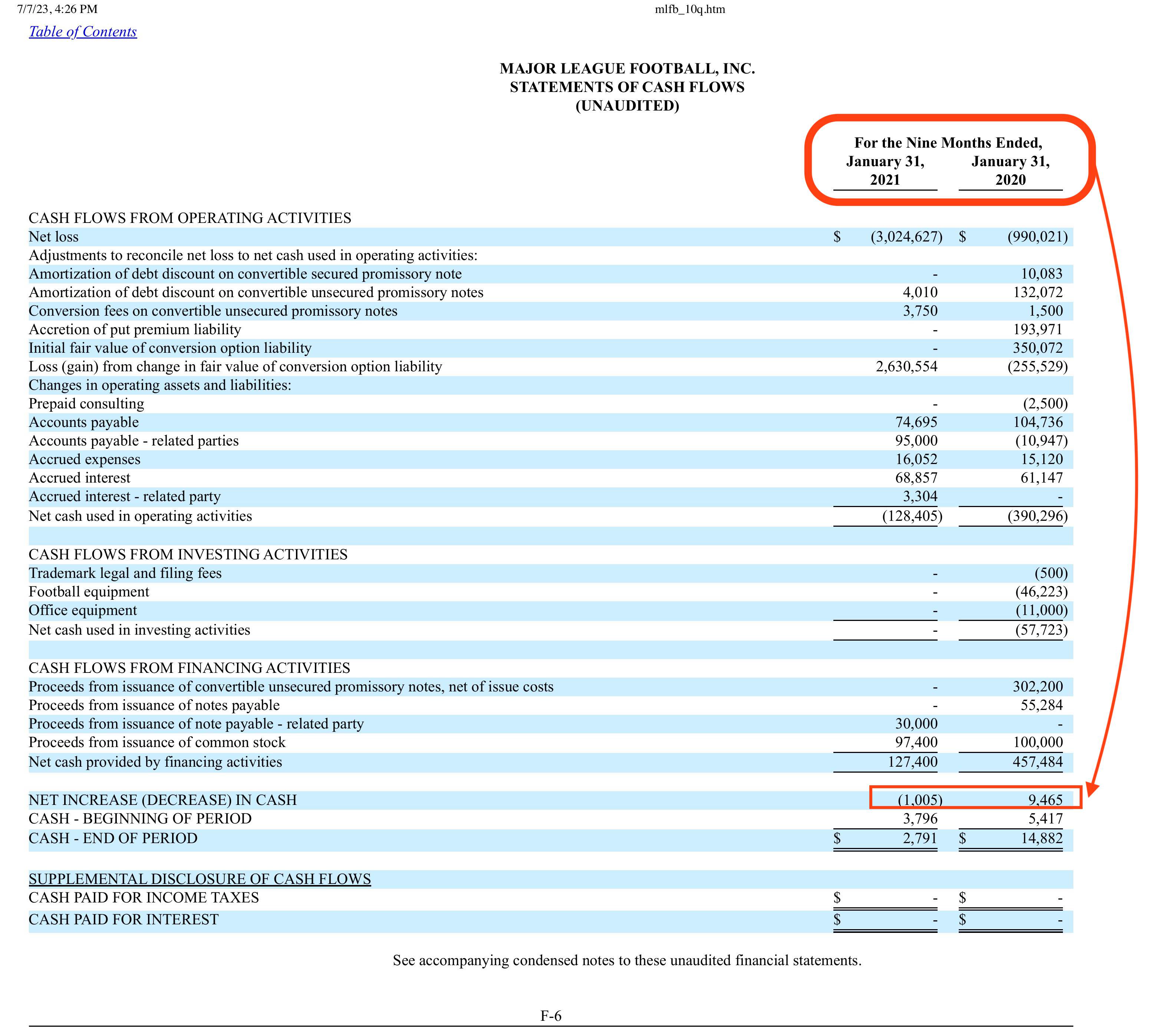}
\caption{\small  Page 6 from an example document from the \dataset\\ test set. Since the original document has 6 pages which can not fit in a single forward pass of 512 tokens, the document is split into several chunks, leading to a loss of important content. For example, in this table from the sixth page, the context from the top is crucial to decide whether to pick the net change in cash entity or not, since we are only interested to extract quarter information "Three months" periods only. }
\vspace{-4mm}
\label{fig:test_example_pred}
\end{figure*}

To provide a deeper analysis of the results on the \dataset\\ dataset, we conduct ablations and report metrics by entity for both LiLT \cite{lilt} and \model\\, as shown in Table \ref{tab:longfin_ablations}. We notice that the gap in performance is more significant in the entities that are typically found in long tables such as Beginning Cash, Ending Cash, Financial Cash, and Change in Cash. 
To illustrate the challenges posed by long tables, we present an examples from our test set in Figure \ref{fig:test_example_pred}. In the example, the table header indicates "Nine Months," implying that the table includes information for a nine-month period that should not be extracted as we are only interested in the financial information per quarter "Three Months". 
Due to the large number of rows and content in the table, the short-context models may not be able to include all the table information in a single forward pass of 512 tokens. Consequently, when the long documents are split into chunks, such tables might be divided as well, leading to the short-context models losing important context when making predictions.

\begin{table}[t!]
\centering
\scalebox{0.48}{\begin{tabular}{l|cccccc}
\toprule
Model  & \makecell{Beginning \\ Cash} & \makecell{Ending \\ Cash} & \makecell{Financial \\ Cash} &  \makecell{Change in \\ Cash} & \makecell{Quarter \\ Keys} & \makecell{Total \\ Assets}\\
\hline \\
$\text{LiLT}_{BASE}$ & 27.39 & 35.89 & 7.40 & 30.55 & 64.93 & \textbf{47.05}  \\
\rowcolor{gray!20} $\text{\model\\}_{BASE}$  \textcolor{blue}{(ours)}  & \textbf{47.61} & \textbf{56.16} & \textbf{15.15} & \textbf{45.94} & \textbf{75.17} & 45.71  \\
\bottomrule
\end{tabular}
}
\vspace{-2mm}
\caption{
    \small Ablations results of LiLT and LongFin on the \dataset\\ dataset by entity.
}
\vspace{-4mm}
\label{tab:longfin_ablations}
\end{table}

\section{Limitations}
Despite the effectiveness of our model, \model\\, on both short and long context document understanding datasets, it has a few limitations. First, \model\\ was trained and evaluated on the English language only. In future, we plan to expand it to support multiple languages. Second, although \model\\ maximum input length (4096 tokens) can accommodate the multi-page documents in the \dataset\\ dataset as well as most our proprietary datasets, it might not accommodate certain financial documents that contain tens of pages. To overcome this limitation, we may consider further expanding the positional embeddings to accomodate 16K tokens similar to the LED model \cite{longformer} or explore utlizing a model architecture that uses relative position embeddings \cite{shaw-etal-2018-self} such as T5 \cite{t5} instead of the absolute position embeddings. Third, due to limited computational resources, we have not explored many different hyperparameters setup. Hence, there might be room for improvement in our model performance. Finally, while our \dataset\\ shed the light on long context understanding challenges which are frequent in the financial industry, it is still limited in size. We encourage the research community to explore this undercharted area of research since it has various commercial applications in many industries such as finance and legal. 

\vspace{-1mm}
\section{Conclusion}
We introduce \model\\, a multimodal document AI model designed to handle lengthy documents. Additionally, we present the \dataset\\ dataset, which aims to replicate real-world challenges in understanding long contexts, specifically in the financial industry. Through our evaluation, we demonstrate the superior performance of \model\\ on the \dataset\\ dataset, which comprises multi-page documents, while achieving comparable results on previous datasets consisting of single-page documents. Moving forward, our plan is to deploy \model\\ after training it on our proprietary datasets in the finance domain. Furthermore, we are working on extending \model\\ to support different languages.

\appendix
\label{sec:appendix}

\begin{table}[t!]
 \setlength\extrarowheight{2pt}
 \centering
 
 \scalebox{0.60}{\begin{tabular}{l|ccc}
  
    \toprule
   \textbf{Experiment} & \textbf{Steps} & \textbf{Learning Rate} & \textbf{Batch Size} \\ 
   \midrule
   
    \multicolumn{4}{c}{\textbf{Finetuning on the \dataset\\ dataset}}  \\ \midrule
   BERT \cite{bert} & 10000 & 2e-5 & 4 \\
   LiLT \cite{lilt} & 8000 & 2e-5 & 4 \\ 
   Longformer \cite{longformer} & 6000 & 2e-5 & 4\\
   \model\\ \textcolor{blue}{(Ours)} & 6000 & 2e-5 & 4 \\ 
   \midrule
   
    \multicolumn{4}{c}{\textbf{Finetuning \model\\ on the previous datasets}}  \\
    \midrule
   $\model\\_{FUNSD}$ & 6000 & 2e-5 & 4 \\ 
   $\model\\_{CORD}$ & 6000 & 5e-5 & 8 \\ 
    \bottomrule

 \end{tabular}}
 
 \caption{
  Training details for finetuning the different models on the \dataset\\ dataset. The lower section also shows the hyperparameters used in finetuning \model\\ on the previous single-page datasets.
 }
 \label{tab:finetuningdetails}
\end{table}

\section{Ethical Statement}
All the documents used in our \dataset\\ dataset is collected from the EDGAR database which grants the right to use and distribute their data without permissions \footnote{https://www.sec.gov/privacy\#dissemination}. The dataset annotation process were accomplished by data annotators who are fairly compensated.  We provide the hyperparameters and experimental setups of our experiments to ensure the reproducibility of our work. Moreover, the models, LiLT \cite{lilt} and Longformer \cite{longformer}, on which our LongFin model is built are published under permissive licenses \footnote{https://github.com/allenai/longformer}\footnote{ https://github.com/jpWang/LiLT} that allow commercial use.

\bibliography{aaai24}

\end{document}